\documentclass{article}

\usepackage{arxiv}

\usepackage[utf8]{inputenc} 
\usepackage[T1]{fontenc}    
\usepackage{hyperref}       
\usepackage{url}            
\usepackage{booktabs}       
\usepackage{amsfonts}       
\usepackage{nicefrac}       
\usepackage{microtype}      
\usepackage{natbib}
\usepackage{amsmath}

\usepackage{subfigure}
\usepackage{graphicx}
\usepackage{multirow}
\hypersetup{
    colorlinks,
    linkcolor=blue,
    filecolor=blue,
    urlcolor=blue,
    citecolor=blue,
}
\DeclareUnicodeCharacter{2113}{$\ell$}

\title{Dual sparse training framework: inducing activation map sparsity via Transformed $\ell1$ regularization}

\author{ 
    Xiaolong Yu\\
    Institute of Computing Theory and Technology\\
	Xidian University\\
    Xi'an, China\\
	\texttt{xl\_yu@stu.xidian.edu.cn} \\
	\And
	Cong Tian\\
    Institute of Computing Theory and Technology\\
	Xidian University\\
    Xi'an, China\\
	\texttt{ctian@mail.xidian.edu.cn} \\
}

\date{}

\renewcommand{\shorttitle}{Dual Sparse Training framework}

\begin{document}

\maketitle

\begin{abstract}
    Although deep convolutional neural networks have achieved rapid development, it is challenging to widely promote and apply these models on low-power devices, due to computational and storage limitations. To address this issue, researchers have proposed techniques such as model compression, activation sparsity induction, and hardware accelerators. This paper presents a method to induce the sparsity of activation maps based on Transformed $\ell1$ regularization, so as to improve the research in the field of activation sparsity induction. Further, the method is innovatively combined with traditional pruning, constituting a dual sparse training framework. Compared to previous methods, Transformed $\ell1$ can achieve higher sparsity\footnote{All sparsity mentioned in this paper refers to the percentage of zero values.} and better adapt to different network structures. Experimental results show that the method achieves improvements by more than 20\% in activation map sparsity on most models and corresponding datasets without compromising the accuracy. Specifically, it achieves a 27.52\% improvement for ResNet18 on the ImageNet dataset, and a 44.04\% improvement for LeNet5 on the MNIST dataset. In addition, the dual sparse training framework can greatly reduce the computational load and provide potential for reducing the required storage during runtime. Specifically, the ResNet18 and ResNet50 models obtained by the dual sparse training framework respectively reduce 81.7\% and 84.13\% of multiplicative floating-point operations, while maintaining accuracy and a low pruning rate.
\end{abstract}


\section{Introduction}
    In recent years, deep neural networks have demonstrated exceptional performance and achieved remarkable success in various fields. For instance, in image classification \citep{krizhevsky2012imagenet, he2016resnet, huang2017densenet}, deep neural networks have significantly improved classification accuracy. In the fields of speech recognition \citep{gulati2020conformer} and natural language processing \citep{brown2020language}, they have also made groundbreaking advancements. These successes often rely on vast model parameters and expensive computational resources. For example, ResNet18 \citep{he2016resnet} includes up to 1.81G FLOPS of computation and 11.7M model parameters. GPT-3 \citep{brown2020language} is of even larger scale with up to 175 billion parameters, and the computation for a single token is six times the number of parameters. These frequent parameter accesses and intensive computations not only result in high energy consumption but also lead to longer inference delays, limiting the application of these models on resource-constrained, low-power devices. Therefore, reducing computational costs and energy consumption while maintaining performance is particularly important.

    A significant amount of research has been dedicated to promoting the effective application of these models on low-power devices through various methods. Neural network pruning \citep{han2015deep, lee2018snip, frankle2018the, ma2019transformed, tanaka2020pruning, tang2021manifold} is one of such techniques, which compresses the network and reduces computational load by decreasing the number of weights (i.e., setting unimportant parameters to zero). Another approach is knowledge distillation \citep{hinton2015distilling, tian2019contrastive}, which extracts smaller dense networks from larger dense networks and transfers the knowledge learned by the large networks to the smaller ones. While these studies focus on increasing the sparsity of model parameters, there is relatively less research on how to increase the sparsity of activation maps. Increasing the sparsity of activation maps can further reduce computational costs and memory usage, making deep neural networks more efficient in resource-constrained environments.

    Most models use the ReLU activation function \citep{glorot2011deep}, which provides natural sparsity to activation maps due to its unique structure. This characteristic is widely utilized in many hardware accelerators \citep{reagen2016minerva, han2016eie, kim2017zena, parashar2017scnn, chen2019eyeriss, wang2021dual}, which reduce computation by skipping over zero values in the activation maps or use sparse representation to save storage space. However, the current level of activation map sparsity is not sufficient to fully unlock the performance of accelerators. To address this issue, a series of optimization methods have been proposed \citep{georgiadis2019accelerating, kurtz2020inducing, zhu2022arts, zhu2023star}, aiming to induce more zero values in the activation maps and thereby further enhance the efficiency of accelerators.

    \begin{figure}[t]
        \centering
        \subfigure[The sparsity of activation maps]{
        \includegraphics[width=0.4\columnwidth]{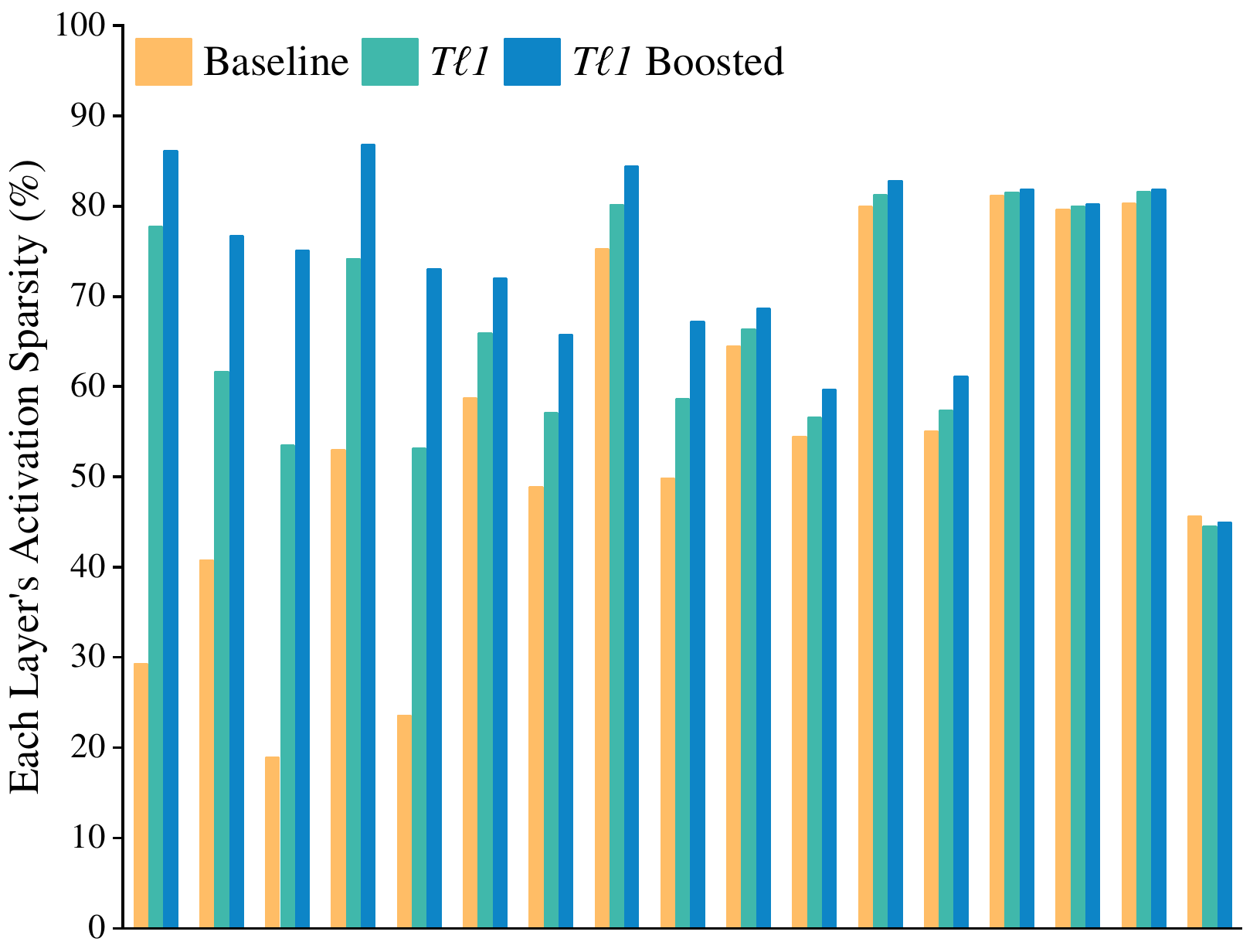}
        }
        \qquad
        \subfigure[The sparsity of model weights]{
        \includegraphics[width=0.4\columnwidth]{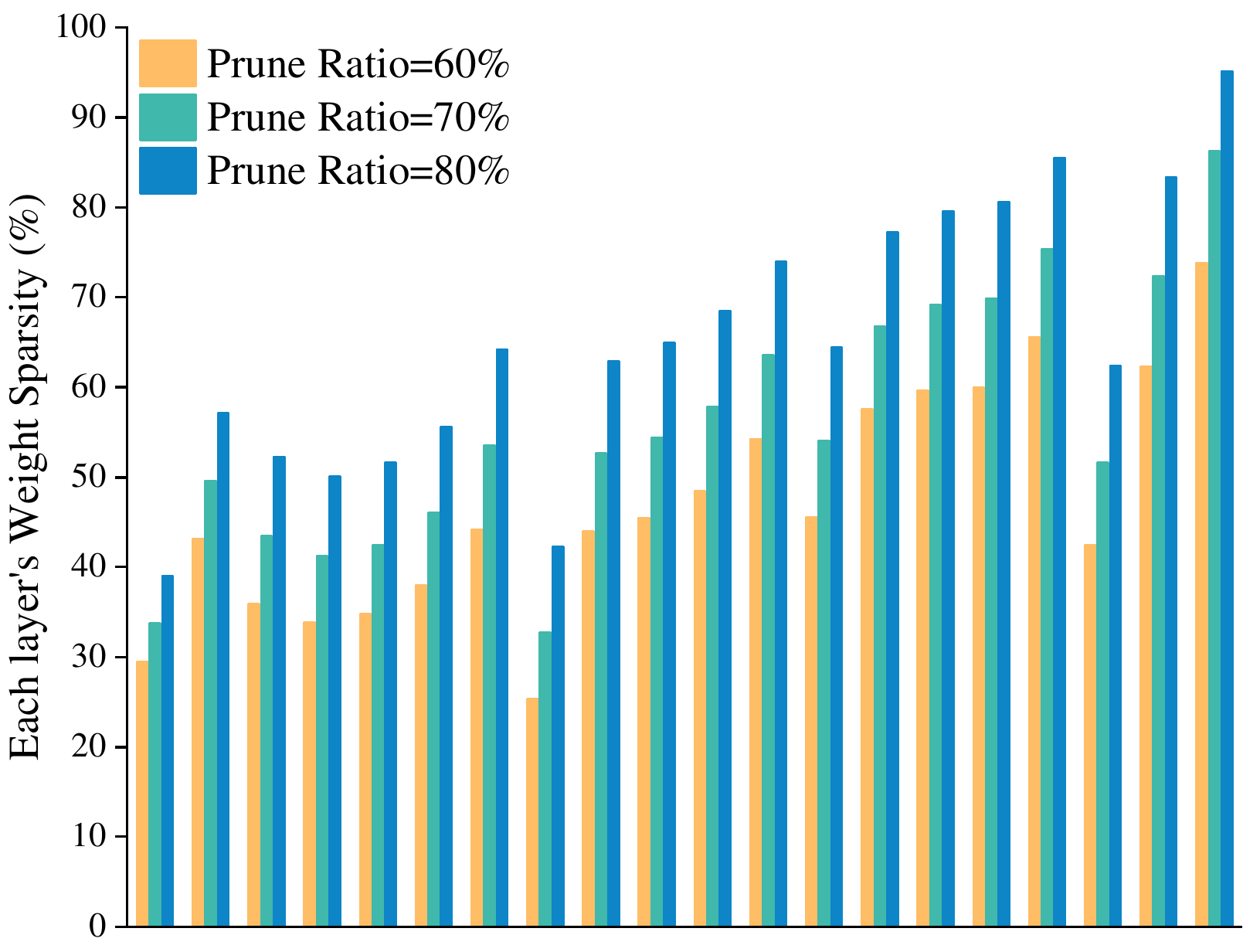}
        }
        \caption{The sparsity of activation maps and weights for each layer of ResNet18} \label{fig:dualSparsity}
    \end{figure}

    Currently, methods based on regularization techniques are widely used in the study of activation map sparsity. However, existing regularizers such as $\ell1$ \citep{georgiadis2019accelerating} and Hoyer \citep{kurtz2020inducing} have certain limitations, including slow convergence, long training periods, insignificant sparsity improvement, and lack of generalizability. These issues restrict their effectiveness and universal applicability in practice. To improve such situation, the first contribution of this paper is the introduction of a method for inducing activation map sparsity based on Transformed $\ell1$ ($T\ell1$). This method has a smoother and faster convergence process suitable for various model architectures, and can significantly enhance activation sparsity across different models. Experimental results show that networks with $T\ell1$ sparsity induction, while maintaining accuracy, achieve an improvement by more than 20\% activation sparsity on most models and datasets. Moreover, its accuracy and the extent of activation sparsity improvement are superior to those of the $\ell1$ and Hoyer based methods.

    Methods to accelerate models can not only increase single-sided sparsity but also combine two types of sparsity for a composite effect. Although several accelerators utilizing both types of sparsity have emerged \citep{kim2017zena, parashar2017scnn, chen2019eyeriss, wang2021dual}, the natural sparsity of activation maps alone cannot fully unlock the performance of these accelerators. Observe that the induction of activation map sparsity primarily increases the sparsity in shallow layers of the network, while pruning mainly improves the sparsity in deeper layers (see Figure \ref{fig:dualSparsity} for an example based on ResNet18). Based on this observation, this paper proposes a novel approach that combines model pruning and activation map sparsity induction, constituting a dual sparse training framework that integrates model pre-training, weight sparsification, and activation map sparsification. The framework can significantly reduce model inference computations with minimal training costs, as the weight-side sparsification employs $\ell1$ regularization threshold pruning. On the ImageNet-1k dataset, the ResNet18/34/50 networks trained under the framework, with a 60\% pruning rate, achieve a computation reduction comparable to that with an 80\% pruning rate achieved by pruning alone.

\section{Related work}
    Numerous researches have been carried out in the field of neural network pruning, see \citep{hoefler2021sparsity, cheng2023survey} for an overview. Magnitude pruning for weight sparsification \citep{han2015deep} is adopted by our work. This method first sets weights with absolute values below a predefined threshold to zero, and then fine-tunes the model to restore its accuracy. Additionally, it integrates quantization and Huffman coding into a three-stage pipeline compression scheme, achieving 35x and 49x compressions for AlexNet and VGG-16 on the ImageNet dataset, respectively. This multi-stage compression strategy not only significantly reduces the model’s storage requirements but also greatly enhances the model’s operational efficiency without significantly impacting accuracy.

    Model pruning offers a solution for static sparsity, whereas previous research on dynamic sparsity, i.e. activation map sparsity, primarily relies on the inherent sparsity of activation maps to construct neural network accelerators. The studies \citep{reagen2016minerva, han2016eie} successfully avoid referencing or computing corresponding weights by leveraging zero values in activation maps, achieving 50\% and 65.16\% energy savings, respectively, but their applicability is limited to accelerating fully connected layers. Then, \citep{kim2017zena, parashar2017scnn, chen2019eyeriss, wang2021dual} not only introduce these techniques to convolutional layers but also exploit weight sparsity. Parashar et al. \citep{parashar2017scnn} explore the performance limits of SCNN by manually adjusting the sparsity of weights and activation maps in various layers of GoogleNet. They found out that when both sparsities are 10\% the performance can be enhanced by 24x, while constrained by actual sparsity they ultimately achieve only a 2.19x performance improvement. This study also experimentally demonstrates that leveraging both types of sparsity simultaneously is superior to using either one alone. The development of these accelerators provides us with the opportunity to construct a dual sparse training framework. Furthermore, some recent work is focused on simultaneously leveraging model-side and activation-side sparsity from an algorithmic perspective. Raihan et al. \citep{raihan2020sparse} use the Top-K concept to retain only the top K magnitudes of weights and activation maps during backpropagation to accelerate model training. The studies \citep{akiva2022searching, mukherji2023activity} demonstrate that the combination of activation and weight pruning is superior to individual weight or activation pruning. Akiva-Hochman et al. \citep{akiva2022searching} use an NAS-based strategy combined with the N:M fine-grained structured sparsity method to select appropriate sparsification directions for each layer of the network.

    Research on how to increase activation map sparsity is relatively limited. Yang et al. \citep{yang2019dasnet} propose increasing activation-side sparsity by setting dynamic activation masks based on neuron importance. However, this method can only achieve limited sparsity and increases computational complexity due to the additional operations required at runtime. Oh et al. \citep{oh2021exploiting} use threshold activation functions in certain layers, setting the smallest K\% of values to zero. This method is then applied to ViT \citep{li2022lazy}. Grimaldi et al. \citep{grimaldi2023accelerating} induce semi-structured activation sparsity through minor runtime modifications. The studies most closely related to this work are \citep{georgiadis2019accelerating, kurtz2020inducing}. The method \citep{georgiadis2019accelerating} improves activation sparsity by adding $\ell1$ regularization on activation maps to the loss function and retraining the network. Based on this method, the frameworks Arts and STAR \citep{zhu2022arts, zhu2023star} are proposed. Arts further increases activation-side sparsity by adaptively changing the penalty coefficient, while STAR combines activation regularization and thresholding to overcome the limitations of using a single regularization or thresholding approach. The method \citep{kurtz2020inducing} uses Hoyer regularization which significantly reduces the time required for activation-side sparsity induction compared to $\ell1$ (90 epochs for the former vs. 20 for the latter). However, the differences in how to improve activation map sparsity among these methods are not substantial, and experiments have shown that Hoyer is not always superior to $\ell1$. This paper proposes a new regularizer, Transformed $\ell1$, to this field, which outperforms existing methods on most models.

\section{Inducing activation map sparsity via Transformed $\ell1$ regularization}
    In deep convolutional neural networks, the commonly used loss functions are of the form:
    \begin{equation}\label{eq:traditonalCostFunction}
        E_0(w)=\frac{1}{N}\sum_{n=1}^{N}C_0(y_n,f_{n,L}(w))
    \end{equation}
    where $N$ is the mini-batch size, $y_n$ is the label corresponding to the $n$-th example, $f_{n,L}(w)$ is the output of the $n$-th sample at the $L$-th layer of the network (typically the last layer), and $C_0(y_n,f_{n,L}(w))$ is the loss function (usually cross-entropy).

    It is a common practice to use regularizers on Equation (\ref{eq:traditonalCostFunction}) to induce weight/activation map sparsity as follows.
    \begin{equation}\label{eq:costFunctionOfActsSparsityInducing}
        \begin{aligned}
            E(w) & = E_0(w)+\frac{1}{N}\sum_{n=1}^{N}\sum_{l=1}^{L-1}\alpha_l Re(f_{n,l}(w))   \\
                 & = \frac{1}{N}\sum_{n=1}^{N}C_n(w)
        \end{aligned}
    \end{equation}
    For activation-side sparsity induction using regularizers, Equation (\ref{eq:costFunctionOfActsSparsityInducing}) is commonly employed. Here, the notation $C_n(w)$ is a shorthand as:
    \begin{equation}
        C_n(w)=C_0(y_n,f_{n,L}(w))+ \sum\limits_{l=1}^{L-1}\alpha_lRe(f_{n,l}(w))
    \end{equation}
    and its gradient during the backpropagation process is calculated as follows.
    \begin{equation}\label{eq:backPropogation}
        \begin{aligned}
            \frac{\partial C_n(w)}{\partial w_i}&=\frac{\partial C_0(y_n,f_{n,L}(w))}{\partial w_i} + \sum_{l=i}^{L-1}\alpha_l\frac{\partial Re(f_{n,l}(w))}{\partial w_i}\\
                &=\frac{\partial C_0(y_n,f_{n,L}(w))}{\partial w_i} + \sum_{l=i}^{L-1}\alpha_l\frac{\partial Re(f_{n,l}(w))}{\partial f_{n,l}(w)}\cdot \frac{\partial f_{n,l}(w)}{\partial w_i}
            \end{aligned}
    \end{equation}
    As is shown in Equation (\ref{eq:backPropogation}), the impact increases as gradients propagate backwardly, which explains why these methods primarily enhance the sparsity of activation maps in the shallow layers.

    The $\ell0$ norm is the most intuitive form of sparse regularization function, as it can achieve the highest degree of sparsity. However, research shows that minimizing the $\ell0$ norm is an NP-hard problem \citep{natarajan1995sparse}, meaning its solution process is complex and computationally expensive. In contrast, the $\ell1$ norm is often regarded as an alternative to $\ell0$ because it is convex and thus easier to solve. However, the $\ell1$ norm is very sensitive to outliers, which may cause oscillations during the training process, making the model difficult to converge. Figure \ref{fig:L1_HS_TL1} illustrates this phenomenon. The square Hoyer regularizer ($Hoyer(\mathbf{x})=\frac{||\mathbf{x}||_1^2}{||\mathbf{x}||_2^2}$) combines the $\ell1$ and $\ell2$ norms to measure sparsity through their ratio, but its optimization process is more complex, requiring more computational resources and time.

    This paper employs the Transformed $\ell1$ ($T\ell1$) regularizer, which is generally defined as follows.
    \begin{equation}
        \label{eq:tranformed_L1}
        T\ell1(x)=\frac{(1+\beta)|x|}{\beta+|x|},\ \beta>0,
    \end{equation}
    When acting on tensors, the regularizer can be formulated in the following way.
    \begin{equation}
        \label{eq:tranformed_L1_tensor}
        T\ell1(\mathbf{x})=\sum_{x_i}\frac{(1+\beta)|x_i|}{\beta+|x_i|},\ \beta>0
    \end{equation}
    By adjusting the value of $\beta$, $T\ell1$ can smoothly interpolate between the $\ell0$ and $\ell1$ norms. 
    \begin{equation}
        \label{eq:tl1_limitation}
        \lim\limits_{\beta\to0^+}T\ell1(\mathbf{x})=||\mathbf{x}||_0\quad and\quad \lim\limits_{\beta\to+\infty}T\ell1(\mathbf{x})=||\mathbf{x}||_1
    \end{equation}

    Moreover, $T\ell1$ satisfies the three conditions of an excellent penalty function \citep{zhang2017transformed}: unbiasedness, sparsity, and Lipschitz continuity \citep{fan2001variable}. The work \citep{ma2019transformed} applies $T\ell1$ to neural network pruning, demonstrating its applicability in neural network sparsification. To the best of our knowledge, this paper is the first to investigate its application to activation sparsity.

\section{Dual sparse training framework}
    \begin{figure}[t]
        \centering
        \includegraphics[scale=0.8]{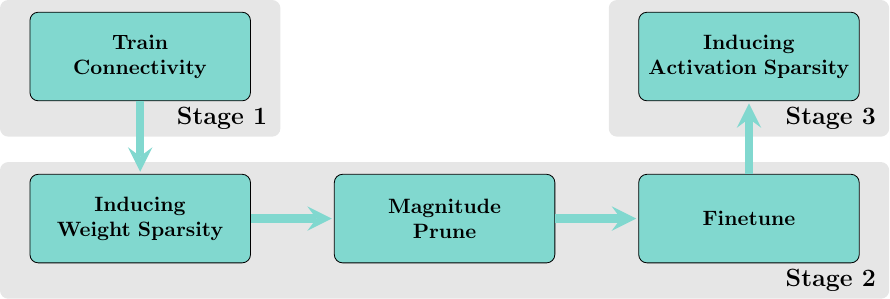}
        \caption{Dual sparse training framework}\label{fig:dstf}
    \end{figure}

    Researchers have developed many excellent hardware accelerators that can simultaneously leverage the sparsity of activation maps and weights to reduce the required storage space or skip unnecessary computations. However, due to the limitations of activation map sparsity, these accelerators often fail to fully realize their performance potential in practical applications. In addition, the induction of activation map sparsity primarily improves the sparsity of activations in the shallow layers of the network, thereby mainly reducing the computational load of these layers. By contrast, pruning primarily optimizes the sparsity in the deeper layers of the network, mainly reducing the computational load of these layers. Based on these observations, we propose a dual sparse training framework that combines model pruning and activation map sparsity induction. Specifically, it is a three-stage pipeline approach as illustrated in Figure \ref{fig:dstf}.

    According to the research \citep{liu2018rethinking}, training a densely connected network is indispensable for constructing an efficient sparse network from scratch. This process ensures that the network has sufficiently learned and captured critical data features and internal correlations before entering the sparsification phase. Therefore, in our framework, the first stage optimizes the network’s connectivity structure through traditional training methods, establishing a solid performance foundation.

    The second stage involves sparsifying the weights and model pruning, as illustrated in Figure \ref{fig:dstf}. This stage can be divided into three sub-steps. First, we apply $\ell1$ regularization to the pretrained model obtained from the first stage to induce weight sparsity, causing unimportant parameters to approach zero. Next, we use the $\ell1$ threshold pruning method to set weights with absolute values below the threshold to zero. Finally, we fine-tune the model to restore its accuracy.

    As will be shown in Table \ref{table:dual}, while pruning effectively reduces the number of parameters, it does not significantly increase the sparsity of activation maps. Therefore, the third stage of the framework uses the $T\ell1$ regularizer to finely adjust the activation maps. This stage compensates for the performance improvements not fully realized by weight pruning, through optimizing the dynamic sparsity of the activation maps.

    In this framework, the first stage ensures the basic connectivity of the model, the second stage compresses the model's parameter space through weight sparsification, and the third stage fine-tunes the activation maps to optimize the model's dynamic sparsity and enhance its performance.

    This three-stage pipeline solution provides a comprehensive approach to constructing sparse models, enabling us to significantly improve both the static and dynamic sparsity of the model while maintaining accuracy, thereby enhancing the model's efficiency and inference speed. Moreover, since the latter two stages focus on different aspects, this framework effectively balances the computational burden across the network layers by reducing the number of activations in shallow layers and the number of weights in deep layers, achieving a balanced reduction in the overall computational load of the network.

\section{Experiments}
     In this section, we carry out experiments to evaluate the proposed framework, mainly concerned with two aspects of performance: (1) using $T\ell1$ regularization to induce sparsity in activation maps, and (2) utilizing dual sparse training to induce sparsity in both model weights and activation maps. Specifically, the experiments are conducted on three different datasets: MNIST, CIFAR100, and ImageNet ILSVRC2012, for six different network models: LeNet5, GoogLeNet, DenseNet121, ResNet18/34/50. These models perform exceptionally well in the field of image classification, and the datasets are classic ones widely used by the research community.

    \begin{table}[ht]
        \caption{The impact of different regularizations on inducing sparsity in model activation maps}\label{table:L1_HS_TL1}
        \centering
        \begin{tabular}{cccccc}
        \toprule
        \textbf{DataSets}          & \textbf{Models}               & \textbf{Virants} & \textbf{Top-1 Acc.} & \textbf{Top-5 Acc.} & \textbf{Acts. Sparsity} \\ \midrule
        \multirow{4}{*}{Mnist}     & \multirow{4}{*}{LeNet5}       & Baseline         & 99.27\%             & -                   & 41.95\%                 \\
                                &                               & $\ell1$               & 99.35\%             & -                   & 76.62\%                 \\
                                &                               & Hoyer            & 99.31\%             & -                   & 76.11\%                 \\
                                &                               & $T\ell1$              & \textbf{99.39\%}    & -                   & \textbf{85.99\%}        \\ \midrule
        \multirow{8}{*}{Cifar100}  & \multirow{4}{*}{GoogLeNet}    & Baseline         & \textbf{76.86\%}    & \textbf{93.63\%}    & 55.44\%                 \\
                                &                               & $\ell1$                & 76.57\%             & 93.61\%             & 66.27\%                 \\
                                &                               & Hoyer            & 76.74\%             & 93.52\%             & 69.01\%                 \\
                                &                               & $T\ell1$              & 76.80\%             & 93.30\%             & \textbf{73.66\%}        \\ \cmidrule{2-6}
                                & \multirow{4}{*}{DesnseNet121} & Baseline         & 79.31\%             & 94.50\%             & 59.15\%                 \\
                                &                               & $\ell1$                & 79.04\%             & 94.54\%             & 65.73\%                 \\
                                &                               & Hoyer            & 78.63\%             & 94.43\%             & 75.58\%                 \\
                                &                               & $T\ell1$              & \textbf{79.35\%}    & \textbf{94.67\%}    & \textbf{79.24\%}        \\ \midrule
        \multirow{12}{*}{ImageNet} & \multirow{4}{*}{ResNet18}     & Baseline         & 69.76\%             & 89.08\%             & 40.78\%                 \\
                                &                               & $\ell1$                & 69.93\%             & 89.22\%             & 61.95\%                 \\
                                &                               & Hoyer            & 70.04\%             & 89.15\%             & 59.12\%                 \\
                                &                               & $T\ell1$              & \textbf{70.04\%}    & \textbf{89.35\%}    & \textbf{68.30\%}        \\ \cmidrule{2-6}
                                & \multirow{4}{*}{ResNet34}     & Baseline         & 73.31\%             & 91.42\%             & 43.55\%                 \\
                                &                               & $\ell1$                & 73.71\%             & 91.38\%             & 58.06\%                 \\
                                &                               & Hoyer            & 73.72\%             & 91.46\%             & 58.57\%                 \\
                                &                               & $T\ell1$              & \textbf{73.80\%}    & \textbf{91.54\%}    & \textbf{64.28\%}        \\ \cmidrule{2-6}
                                & \multirow{4}{*}{ResNet50}     & Baseline         & 76.13\%             & 92.86\%             & 44.48\%                 \\
                                &                               & $\ell1$                & 76.53\%             & 93.04\%             & 66.27\%                 \\
                                &                               & Hoyer            & 76.42\%             & 92.96\%             & 63.53\%                 \\
                                &                               & $T\ell1$              & \textbf{76.63\%}    & \textbf{93.11\%}    & \textbf{68.70\%}        \\
        \bottomrule
        \end{tabular}
    \end{table}

    Table \ref{table:L1_HS_TL1} summarizes the impact of our method compared to $\ell1$ and square Hoyer on sparsity. The baseline models for LeNet-5\footnote{https://github.com/ChawDoe/LeNet5-MNIST-PyTorch}, GoogLeNet, and DenseNet121\footnote{https://github.com/weiaicunzai/pytorch-cifar100} are trained from scratch. The ResNet18/34/50 models are obtained from the Pytorch repository\footnote{https://pytorch.org/vision/stable/models.html}. As is shown in Table \ref{table:L1_HS_TL1}, the $T\ell1$ method outperforms both $\ell1$ and Hoyer across various models and datasets, while $\ell1$ and square Hoyer do not always outperform each other. Except for GoogLeNet, under the premise that the accuracy is not compromised or lower than existing methods, the $T\ell1$ regularization achieves the highest sparsity improvements: over 20\% for most models. The sparsity of the activation maps of ResNet18 can be increased by up to 27.52\%, which is 9.18\% higher than square Hoyer and improves the accuracy by 0.28\% compared to the baseline. ResNet50 sees an accuracy improvement of 0.5\% with a 24.22\% increase in activation map sparsity. On the MNIST dataset, the sparsity of LeNet-5 is more than doubled. To determine the hyperparameters $\alpha$ and $\beta$, we use the grid search method that applies the same parameters for each layer, and the final selected parameters are shared in the supplementary materials. The sparse models of ResNet18/34/50 are all fine-tuned from the baseline, with the epoch count for training all the three methods set to 20 (compared to 90 epochs suggested by \citep{georgiadis2019accelerating}). Other training parameters for $\ell1$ and square Hoyer are used as provided in \citep{georgiadis2019accelerating, kurtz2020inducing}.

    \begin{figure}[ht]
        \centering
        \subfigure[LeNet5]{
            \includegraphics[width=0.32\textwidth]{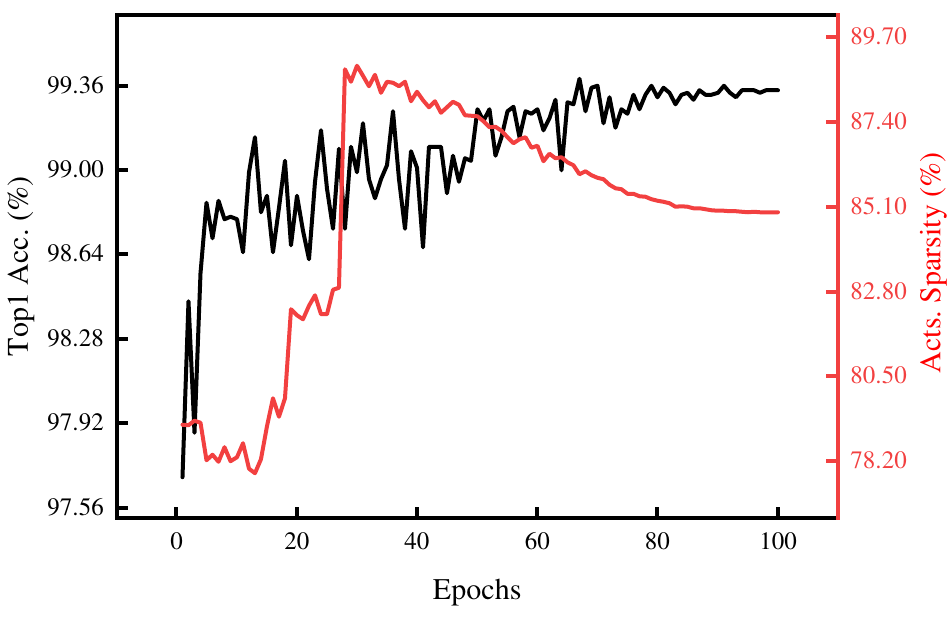}
            \includegraphics[width=0.32\textwidth]{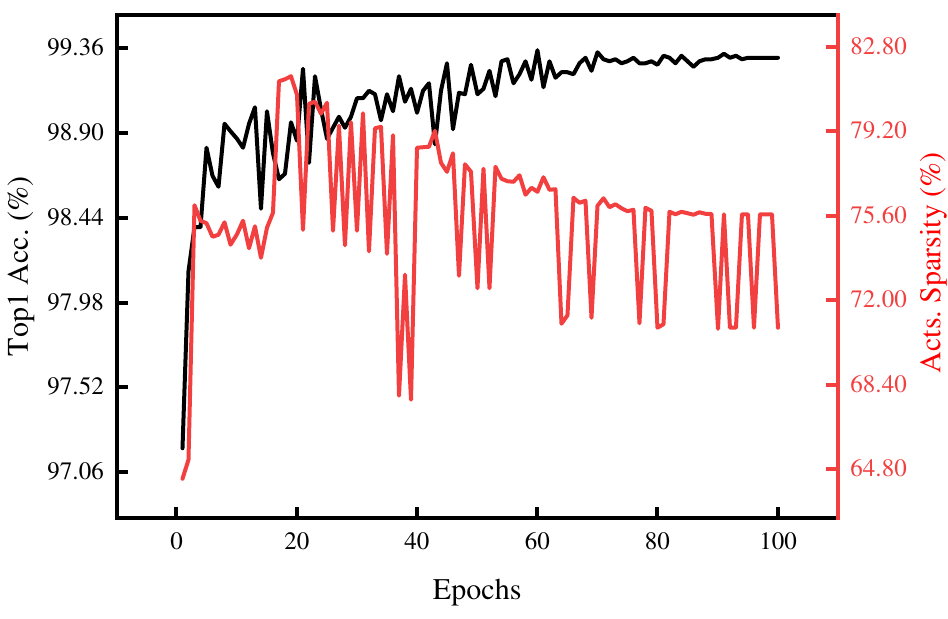}
            \includegraphics[width=0.32\textwidth]{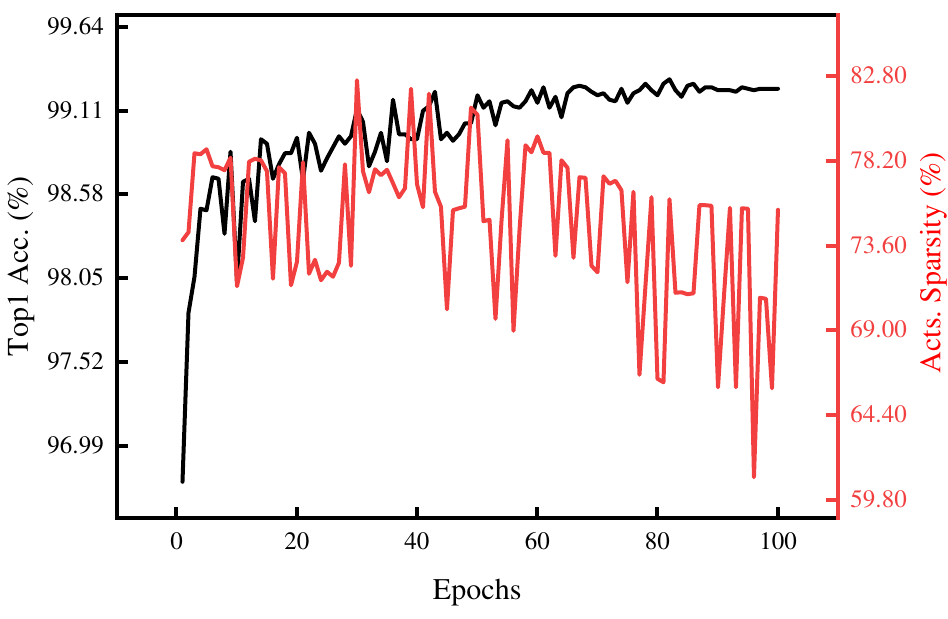}
        }

        \subfigure[DenseNet121]{
            \includegraphics[width=0.32\textwidth]{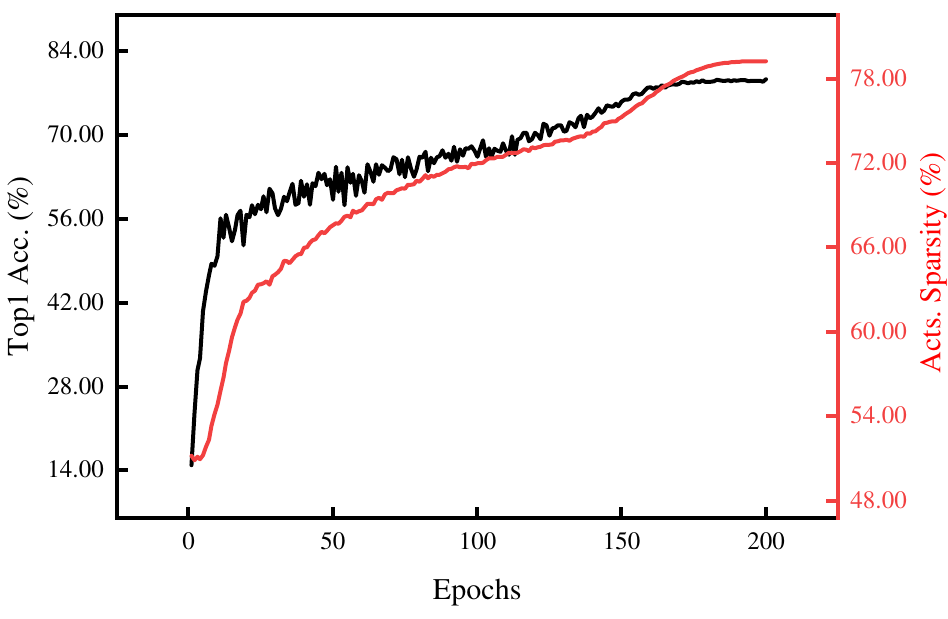}
            \includegraphics[width=0.32\textwidth]{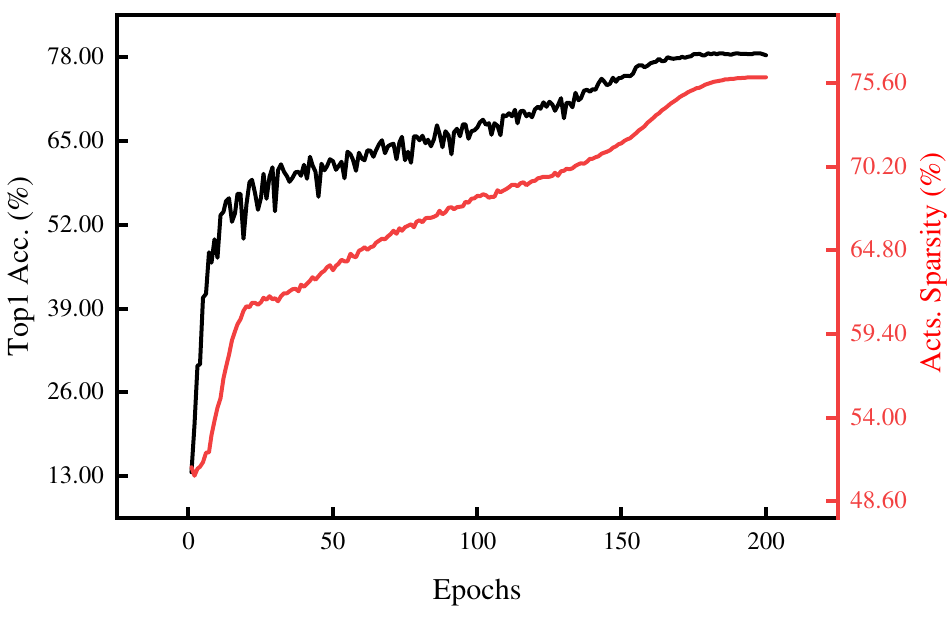}
            \includegraphics[width=0.32\textwidth]{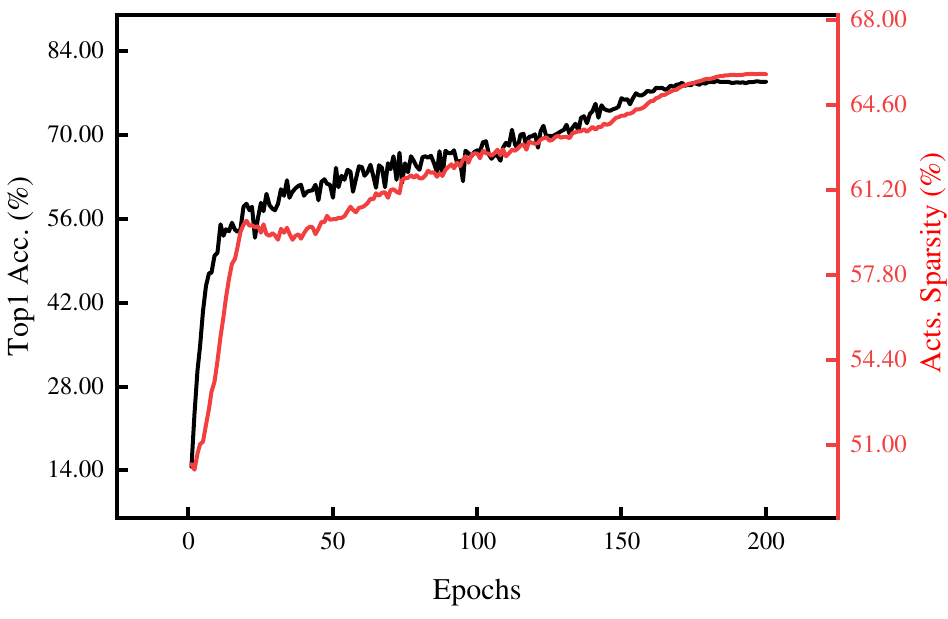}
        }
        \subfigure[ResNet34]{
            \includegraphics[width=0.32\textwidth]{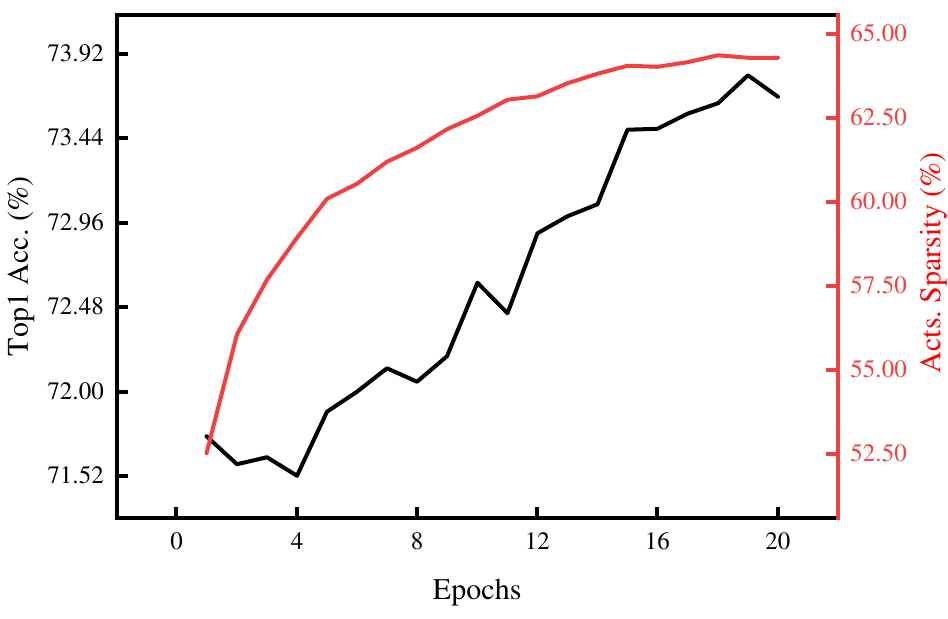}
            \includegraphics[width=0.32\textwidth]{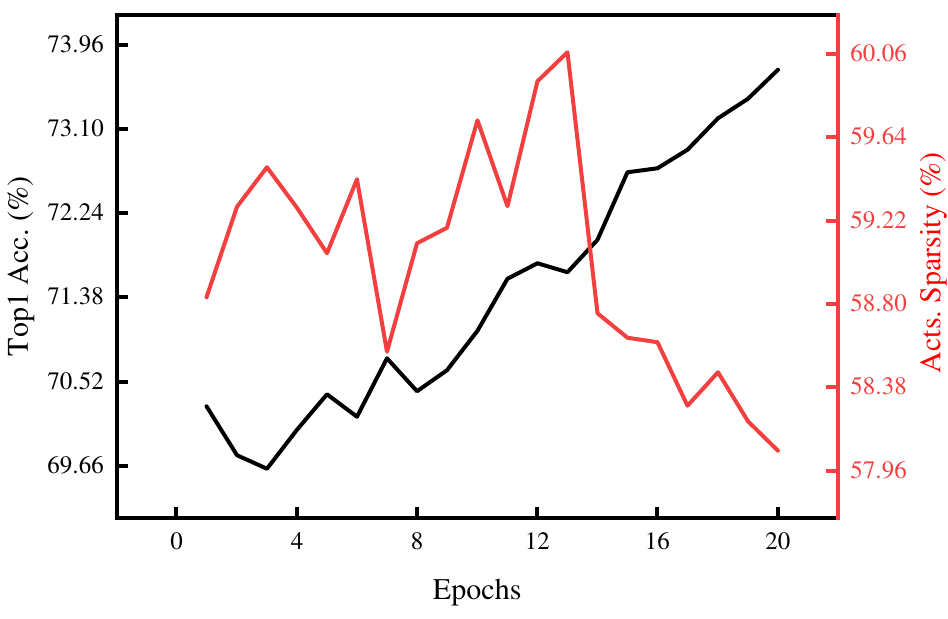}
            \includegraphics[width=0.32\textwidth]{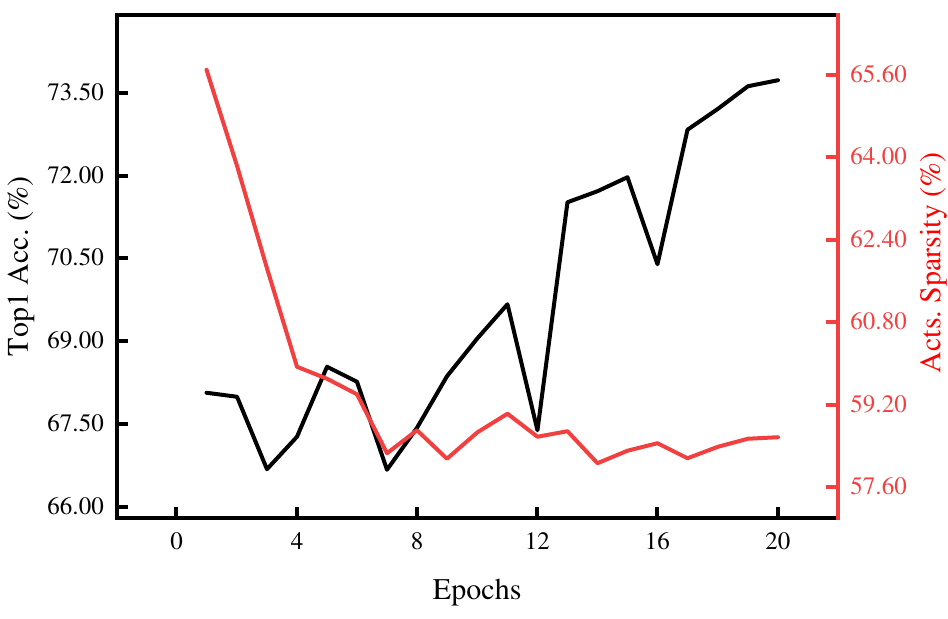}
        }
        \subfigure[ResNet50]{
            \includegraphics[width=0.32\textwidth]{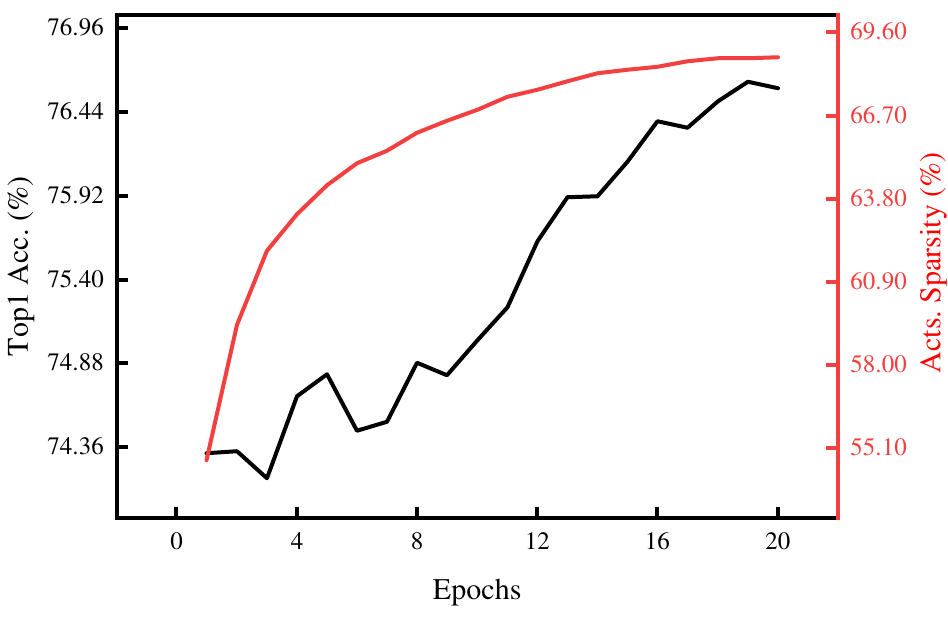}
            \includegraphics[width=0.32\textwidth]{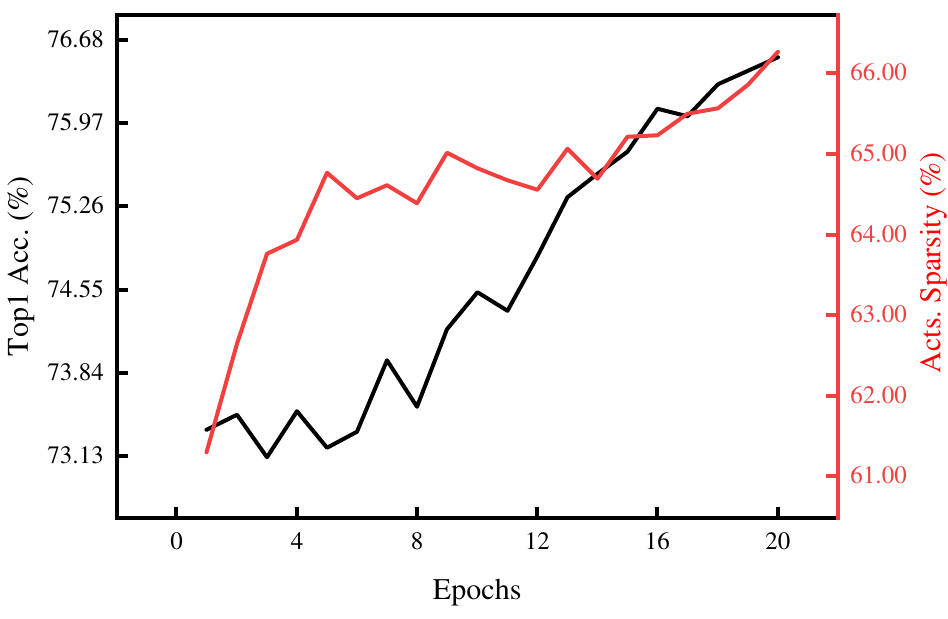}
            \includegraphics[width=0.32\textwidth]{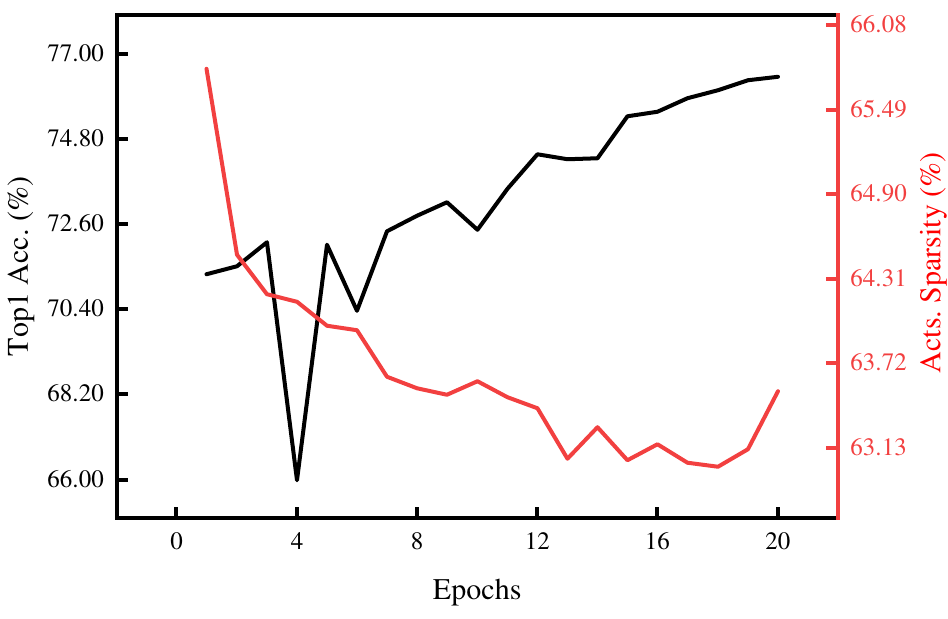}
        }
        \caption{The performance of $T\ell1$ (left), $\ell1$ (center), and square Hoyer (right) regularizers on LeNet5, DenseNet121, ResNet34, and ResNet50, respectively.}\label{fig:L1_HS_TL1}
    \end{figure}

    Figure \ref{fig:L1_HS_TL1} illustrates the changes in Top-1 accuracy and activation map sparsity during the training process when evaluating LeNet-5, DenseNet-121, and ResNet-34/50 on the MNIST, CIFAR-100, and ImageNet validation sets, respectively. As is shown in Figure \ref{fig:L1_HS_TL1}, for the LeNet-5 model, after a slight fluctuation in the initial training stages, $T\ell1$ regularization rapidly increases the sparsity to its peak level, followed by a gradual decmidrule, while the model accuracy steadily converges to its maximum value. In the more complex DenseNet121 and ResNet34/50 models, $T\ell1$ demonstrates more stable and consistent performance. In these models, the sparsity of the activation maps steadily increases along with the training progresses, and the model performance also shows stable recovery or improvement. In contrast, $\ell1$ and square Hoyer regularizers exhibit significant oscillations in three of the models, except for DenseNet121. These oscillations indicate that while trying to enhance sparsity, they struggle to maintain stable model performance, and thus their potential for sparsity enhancement is limited. Overall, compared to $\ell1$ and square Hoyer, $T\ell1$ demonstrates superior regulatory capability across different network architectures and training stages, effectively balancing activation sparsity and network performance.

    To explore the potential mechanisms by which $T\ell1$ steadily enhances activation map sparsity while maintaining model performance, we conduct a detailed quantitative analysis of the parameter changes between the fine-tuned ResNet18/34/50 models and their baseline versions. Euclidean distance and cosine similarity are used as the evaluation metrics. As is shown in Table \ref{table:distance}, compared to the other two regularizers, the models fine-tuned with the $T\ell1$ regularizer exhibit the smallest Euclidean distance and the highest cosine similarity to their baseline counterparts. This indicates that $T\ell1$, in the process of adjusting network parameters to enhance sparsity, retains more of the original model’s information, avoiding substantial alterations to critical parameters that impact model performance. This finding reveals that the effectiveness of $T\ell1$ regularization in promoting activation sparsity while preserving the original model performance lies in its ability to maintain the integrity of crucial parameters.

\begin{figure}[ht]
        \centering
        \subfigure[GoogLeNet]{
        \includegraphics[width=0.4\columnwidth]{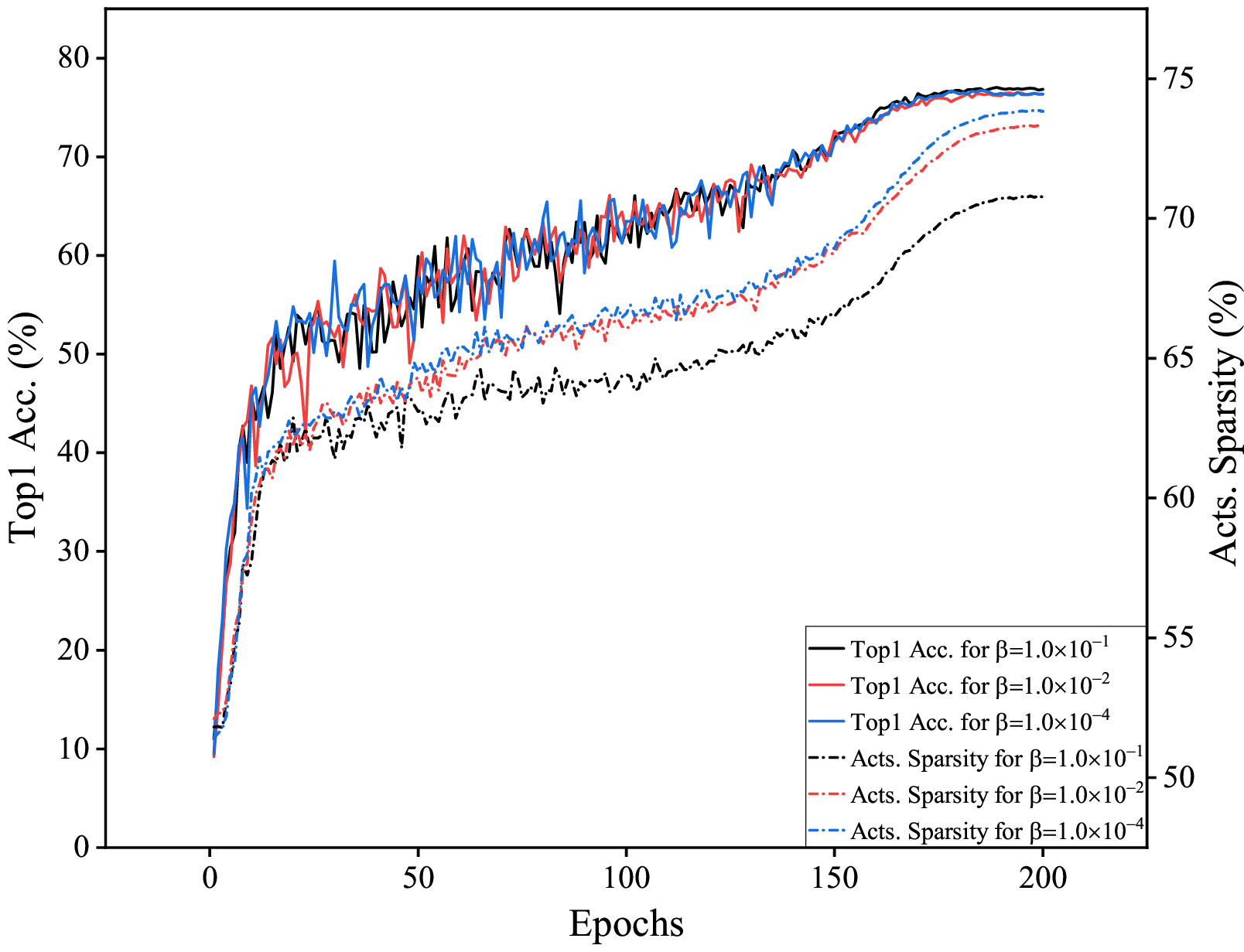}
        }
        \qquad
        \subfigure[ResNet18]{
        \includegraphics[width=0.4\columnwidth]{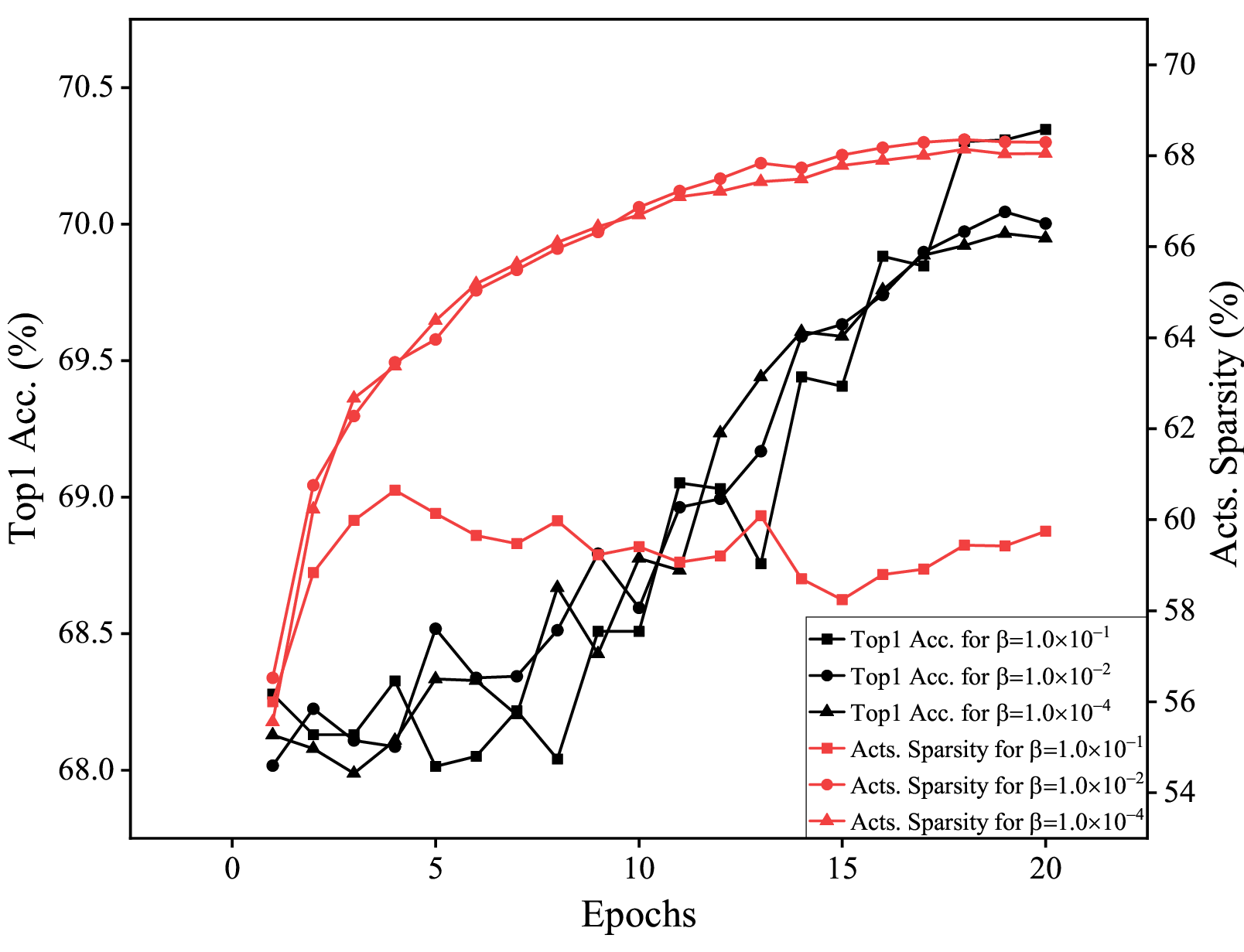}
        }
        \caption{The impact of different values of $\beta$ on Top-1 accuracy and activation map sparsity}\label{fig:versus_beta}
    
    \end{figure}

    We set three different $\beta$ values, specifically $1.0 \times 10^{-1}$, $1.0 \times 10^{-2}$, and $1.0 \times 10^{-4}$, to observe the impact of the $T\ell1$ regularizer on model performance and activation map sparsity. Figure \ref{fig:versus_beta} shows the changes in Top-1 accuracy and activation map sparsity for GoogLeNet and ResNet18 on the CIFAR-100 and ImageNet validation sets, respectively. It can be seen that when $\beta \leq 1.0 \times 10^{-2}$, changes in its value have little effect on the models. However, when $\beta = 1.0 \times 10^{-1}$, the increase in activation map sparsity is reduced, and significant oscillations appear in ResNet18. This phenomenon is related to the working mechanism of $T\ell1$. When $\beta$ is small, $T\ell1$ tends to approximate the optimal mode of sparse representation, i.e. $\ell0$, while with larger values, its behavior degrades to that of $\ell1$.  

    \begin{table}[ht]
        \caption{The Euclidean distance and cosine similarity between ResNet18/34/50 obtained via $T\ell1$, $\ell1$, and square Hoyer regularizations and their respective Baseline}\label{table:distance}
        \centering
        \begin{tabular}{ccccccc}
        \toprule
        \textbf{Model} & ED-$T\ell1$  & ED-$\ell1$  & ED-HS  & CS-$T\ell1$  & CS-$\ell1$     & CS-HS\\
        \midrule
        ResNet18       & \textbf{1.36}         & 1.97        & 1.74   & \textbf{0.96}          & 0.93           & 0.90          \\
        ResNet34       & \textbf{1.12}         & 1.56        & 1.63   & \textbf{0.96}          & 0.94           & 0.91          \\
        ResNet50       & \textbf{1.02}         & 1.38        & 1.63   & \textbf{0.95}          & 0.92           & 0.88          \\
        \bottomrule
        \end{tabular}
    \end{table}

    Following the guidance of Figure \ref{fig:dstf}, we conduct dual sparse training experiments on ResNet18/34/50 using the ImageNet dataset. In the first phase of the experiment, we directly utilize the baseline models obtained from the PyTorch repository. In the second phase, we perform weight sparsification over a total of 40 epochs, including 20 epochs of weight sparsity induction followed by 20 epochs of post-pruning fine-tuning. The third phase involves 20 epochs of activation map sparsification. As is shown in Table \ref{table:dual}, the networks obtained through the dual sparse training framework, under a 60\% pruning rate, significantly reduce more than 81\% of the multiplicative floating-point operations (FLOPs), while the accuracy of the models is not only uncompromised but even slightly improved. For example, the Top-1 accuracy of ResNet34 increases by 0.25\%. In contrast, pure pruning models with an 80\% pruning rate, although slightly advantageous in reducing FLOPs, suffer significant accuracy loss. For instance, the pruned ResNet18 sees an accuracy drop of 0.98\%, and ResNet50 experiences a 0.67\% accuracy decmidrule. It is well known that in many systems with high security requirements, such accuracy loss is intolerable.

    \begin{table}[htbp]
        \caption{Comparison between model pruning and dual sparse training on ResNet18/34/50. WS and AS represent the percentage of zeros in weights and activation maps, respectively. FLOPs Drops refers to the percentage of multiplications resulting in zero in the computational operations.}\label{table:dual}
        \centering
        \begin{tabular}{l|cc|cc|c}
        \multicolumn{1}{c|}{\textbf{Models}}   & \multicolumn{1}{l}{\textbf{Top-1 Acc.}} & \multicolumn{1}{l|}{\textbf{Top-5 Acc.}} & \multicolumn{1}{c}{\textbf{WS}} & \multicolumn{1}{c|}{\textbf{AS}} & \multicolumn{1}{l}{\textbf{Flops Drops}} \\ \midrule
        ResNet18 Baseline & 69.76\%                                 & 89.08\%                                  & -                                         & 40.78\%                                      & 55.71\%                                  \\
        ResNet18 Pruned   & 68.79\%                                 & 88.60\%                                  & \textbf{80.00\%}                             & 40.42\%                                      & 81.22\%                                  \\
        ResNet18 Dual     & \textbf{69.79\%}                        & \textbf{89.18\%}                         & 60.00\%                                      & \textbf{67.82\%}                             & \textbf{81.70\%}                         \\ \midrule
        ResNet34 Baseline & 73.31\%                                 & 91.42\%                                  & -                                         & 43.55\%                                      & 57.42\%                                  \\
        ResNet34 Pruned   & 73.29\%                                 & 91.49\%                                  & \textbf{80.00\%}                          & 43.05\%                                      & \textbf{84.44\%}                         \\
        ResNet34 Dual     & \textbf{73.56\%}                        & \textbf{91.50\%}                         & 60.00\%                                   & \textbf{64.13\%}                             & 81.45\%                                  \\ \midrule
        ResNet50 Baseline & 76.13\%                                 & 92.86\%                                  & -                                         & 44.48\%                                      & 55.09\%                                  \\
        ResNet50 Pruned   & 75.46\%                                 & 92.61\%                                  & \textbf{80.00\%}                          & 43.43\%                                      & \textbf{86.21\%}                         \\
        ResNet50 Dual     & \textbf{76.19\%}                        & \textbf{92.99\%}                         & 60.00\%                                   & \textbf{70.34\%}                             & 84.13\%
        \end{tabular}
    \end{table}

\begin{figure}[ht]
        \centering
        \subfigure[ResNet18]{
        \includegraphics[width=0.4\columnwidth]{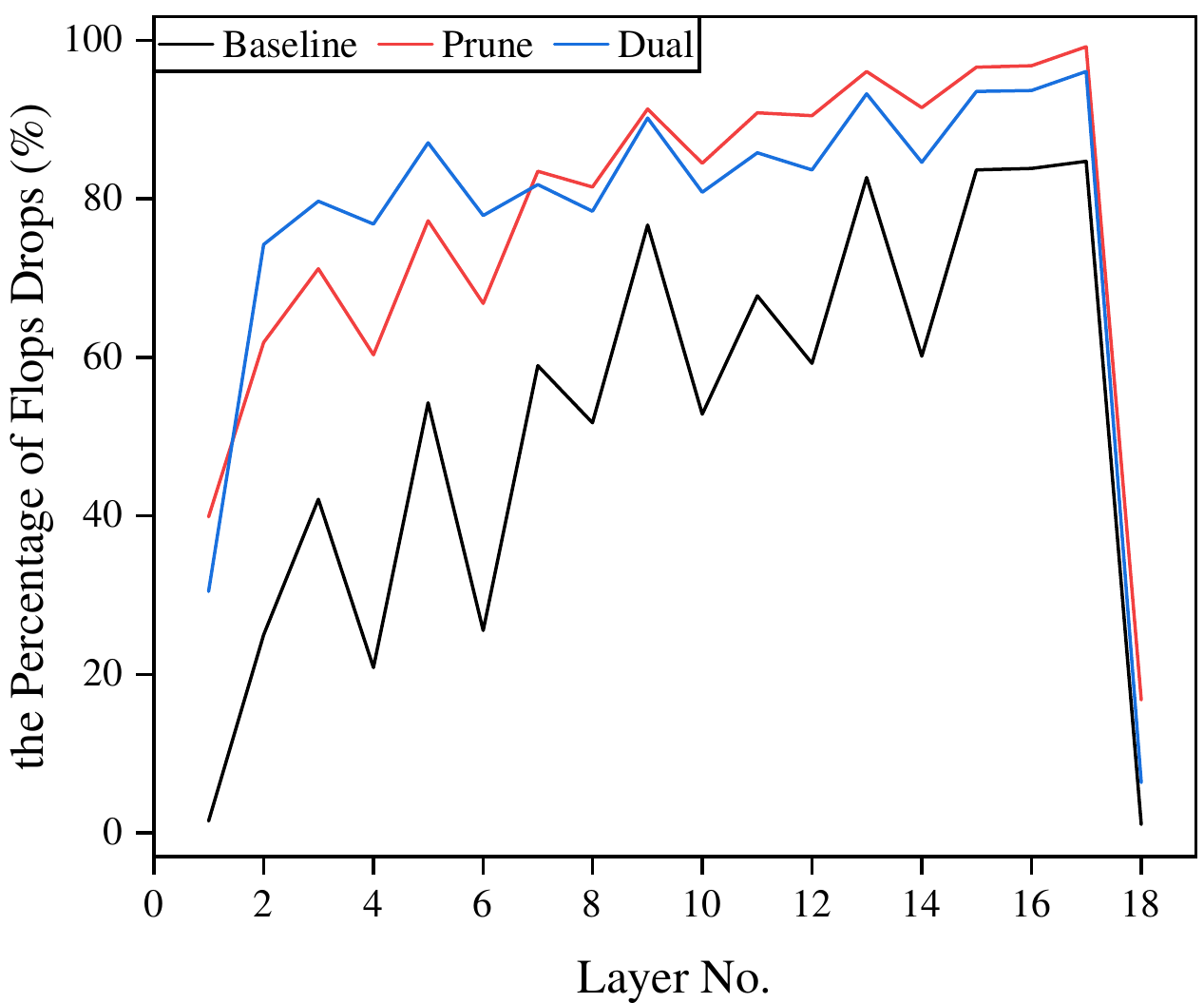}
        }
        \qquad
        \subfigure[ResNet50]{
        \includegraphics[width=0.4\columnwidth]{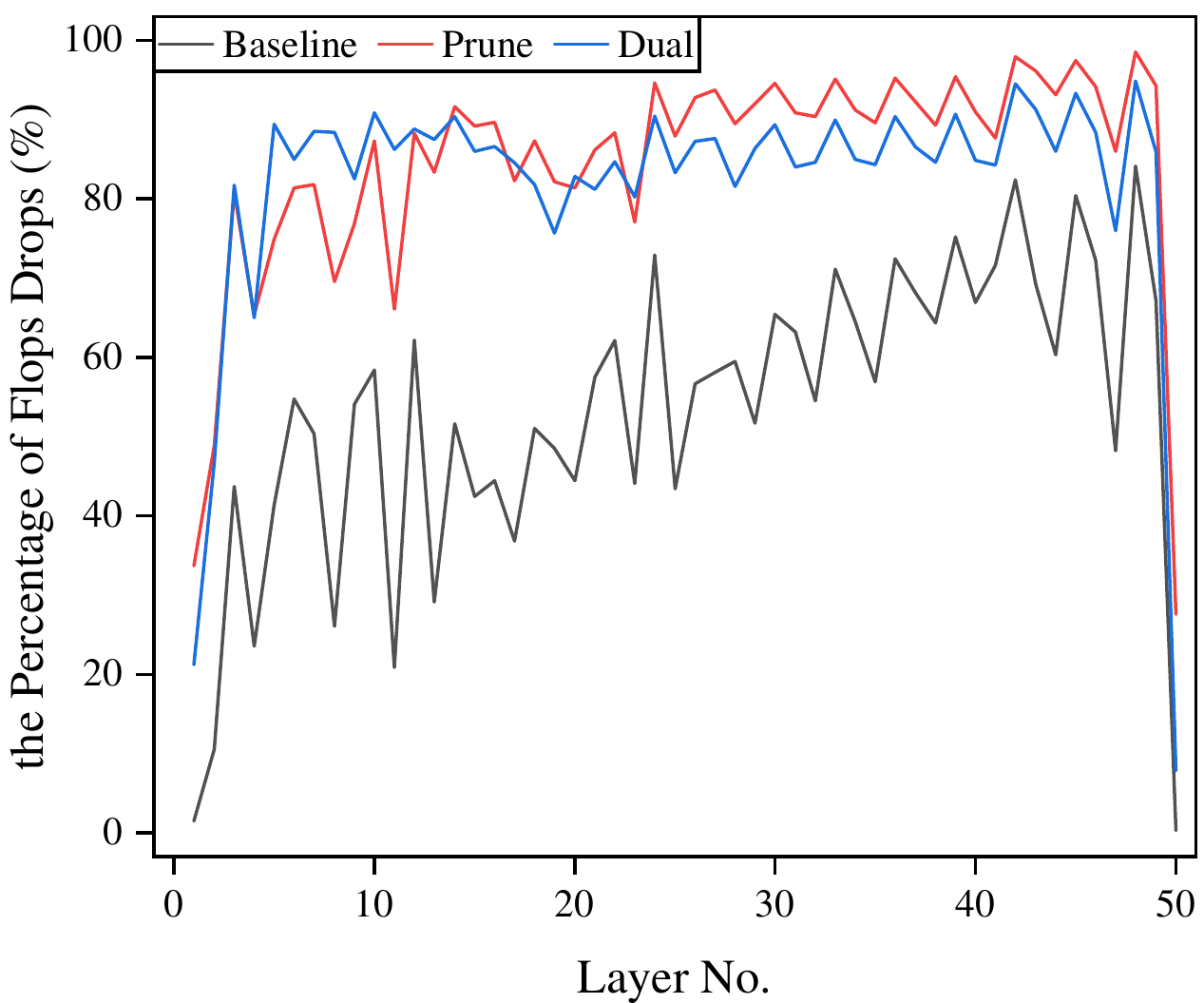}
        }
        \caption{The percentage of Flops Drops per layer for ResNet18/50}\label{fig:flops}
    \end{figure}

    Figure \ref{fig:flops} compiles the performance of ResNet18/50 in terms of FLOPs reduction across various layers detailed in Table \ref{table:dual}. It can be observed that both pure model pruning and dual sparse training result in relatively limited FLOPs reduction in the initial and final layers of the network. This may be because the input and output layers of the network are typically designed to be simpler, making it challenging to achieve substantial computational reductions through pruning or sparsification. More notably, the models obtained through dual sparse training maintain a consistently high and balanced level of FLOPs reduction across the intermediate layers. This suggests that dual sparse training balances the computational load across the network’s layers more effectively, offering valuable insights for optimizing the overall allocation of computational resources within the network. In contrast, the pure pruning models exhibit greater fluctuations in FLOPs reduction across different layers.

\section{Conclusion}
    In this work, we introduce a novel activation map sparsity induction method based on regularization technology, namely Transformed $\ell1$. We validate the performance of this regularizer on multiple public datasets, and the results show that it not only increases the sparsity of activation maps significantly but also enhances the generalization ability of the model. Subsequently, we combine the method with model pruning, formulating a dual sparse training framework. The framework can simultaneously increase the dynamic sparsity generated during model inference and the static sparsity of the model itself, while ensuring that model accuracy is not compromised. In addition, The framework can significantly reduce the memory capacity and bandwidth requirements for real-time image processing, making models easier to deploy on mobile systems, and help accelerators to achieve their optimal performance, so as to improve the overall computational efficiency.

\bibliography{reference}

\begin{thebibliography}{39}
\providecommand{\natexlab}[1]{#1}
\providecommand{\url}[1]{\texttt{#1}}
\expandafter\ifx\csname urlstyle\endcsname\relax
  \providecommand{\doi}[1]{doi: #1}\else
  \providecommand{\doi}{doi: \begingroup \urlstyle{rm}\Url}\fi

\bibitem[Akiva-Hochman et~al.(2022)Akiva-Hochman, Finder, Turek, and Treister]{akiva2022searching}
Ruth Akiva-Hochman, Shahaf~E Finder, Javier~S Turek, and Eran Treister.
\newblock Searching for n: M fine-grained sparsity of weights and activations in neural networks.
\newblock In \emph{European Conference on Computer Vision}, pages 130--143. Springer, 2022.

\bibitem[Brown et~al.(2020)Brown, Mann, Ryder, Subbiah, Kaplan, Dhariwal, Neelakantan, Shyam, Sastry, Askell, et~al.]{brown2020language}
Tom Brown, Benjamin Mann, Nick Ryder, Melanie Subbiah, Jared~D Kaplan, Prafulla Dhariwal, Arvind Neelakantan, Pranav Shyam, Girish Sastry, Amanda Askell, et~al.
\newblock Language models are few-shot learners.
\newblock \emph{Advances in neural information processing systems}, 33:\penalty0 1877--1901, 2020.

\bibitem[Chen et~al.(2019)Chen, Yang, Emer, and Sze]{chen2019eyeriss}
Yu-Hsin Chen, Tien-Ju Yang, Joel Emer, and Vivienne Sze.
\newblock Eyeriss v2: A flexible accelerator for emerging deep neural networks on mobile devices.
\newblock \emph{IEEE Journal on Emerging and Selected Topics in Circuits and Systems}, 9\penalty0 (2):\penalty0 292--308, 2019.

\bibitem[Cheng et~al.(2023)Cheng, Zhang, and Shi]{cheng2023survey}
Hongrong Cheng, Miao Zhang, and Javen~Qinfeng Shi.
\newblock A survey on deep neural network pruning-taxonomy, comparison, analysis, and recommendations.
\newblock \emph{arXiv preprint arXiv:2308.06767}, 2023.

\bibitem[Fan and Li(2001)]{fan2001variable}
Jianqing Fan and Runze Li.
\newblock Variable selection via nonconcave penalized likelihood and its oracle properties.
\newblock \emph{Journal of the American statistical Association}, 96\penalty0 (456):\penalty0 1348--1360, 2001.

\bibitem[Frankle and Carbin(2019)]{frankle2018the}
Jonathan Frankle and Michael Carbin.
\newblock The lottery ticket hypothesis: Finding sparse, trainable neural networks.
\newblock In \emph{International Conference on Learning Representations}, 2019.
\newblock URL \url{https://openreview.net/forum?id=rJl-b3RcF7}.

\bibitem[Georgiadis(2019)]{georgiadis2019accelerating}
Georgios Georgiadis.
\newblock Accelerating convolutional neural networks via activation map compression.
\newblock In \emph{Proceedings of the IEEE/CVF Conference on Computer Vision and Pattern Recognition}, pages 7085--7095, 2019.

\bibitem[Glorot et~al.(2011)Glorot, Bordes, and Bengio]{glorot2011deep}
Xavier Glorot, Antoine Bordes, and Yoshua Bengio.
\newblock Deep sparse rectifier neural networks.
\newblock In \emph{Proceedings of the fourteenth international conference on artificial intelligence and statistics}, pages 315--323. JMLR Workshop and Conference Proceedings, 2011.

\bibitem[Grimaldi et~al.(2023)Grimaldi, Ganji, Lazarevich, and Sah]{grimaldi2023accelerating}
Matteo Grimaldi, Darshan~C Ganji, Ivan Lazarevich, and Sudhakar Sah.
\newblock Accelerating deep neural networks via semi-structured activation sparsity.
\newblock In \emph{Proceedings of the IEEE/CVF International Conference on Computer Vision}, pages 1179--1188, 2023.

\bibitem[Gulati et~al.(2020)Gulati, Qin, Chiu, Parmar, Zhang, Yu, Han, Wang, Zhang, Wu, et~al.]{gulati2020conformer}
Anmol Gulati, James Qin, Chung-Cheng Chiu, Niki Parmar, Yu~Zhang, Jiahui Yu, Wei Han, Shibo Wang, Zhengdong Zhang, Yonghui Wu, et~al.
\newblock Conformer: Convolution-augmented transformer for speech recognition.
\newblock \emph{arXiv preprint arXiv:2005.08100}, 2020.

\bibitem[Han et~al.(2016{\natexlab{a}})Han, Liu, Mao, Pu, Pedram, Horowitz, and Dally]{han2016eie}
Song Han, Xingyu Liu, Huizi Mao, Jing Pu, Ardavan Pedram, Mark~A Horowitz, and William~J Dally.
\newblock Eie: Efficient inference engine on compressed deep neural network.
\newblock \emph{ACM SIGARCH Computer Architecture News}, 44\penalty0 (3):\penalty0 243--254, 2016{\natexlab{a}}.

\bibitem[Han et~al.(2016{\natexlab{b}})Han, Mao, and Dally]{han2015deep}
Song Han, Huizi Mao, and William~J Dally.
\newblock Deep compression: Compressing deep neural networks with pruning, trained quantization and huffman coding.
\newblock \emph{ICLR}, 2016{\natexlab{b}}.

\bibitem[He et~al.(2016)He, Zhang, Ren, and Sun]{he2016resnet}
Kaiming He, Xiangyu Zhang, Shaoqing Ren, and Jian Sun.
\newblock Deep residual learning for image recognition.
\newblock In \emph{Proceedings of the IEEE conference on computer vision and pattern recognition}, pages 770--778, 2016.

\bibitem[Hinton et~al.(2015)Hinton, Vinyals, and Dean]{hinton2015distilling}
Geoffrey Hinton, Oriol Vinyals, and Jeffrey Dean.
\newblock Distilling the knowledge in a neural network.
\newblock In \emph{NIPS Deep Learning and Representation Learning Workshop}, 2015.
\newblock URL \url{http://arxiv.org/abs/1503.02531}.

\bibitem[Hoefler et~al.(2021)Hoefler, Alistarh, Ben-Nun, Dryden, and Peste]{hoefler2021sparsity}
Torsten Hoefler, Dan Alistarh, Tal Ben-Nun, Nikoli Dryden, and Alexandra Peste.
\newblock Sparsity in deep learning: Pruning and growth for efficient inference and training in neural networks.
\newblock \emph{Journal of Machine Learning Research}, 22\penalty0 (241):\penalty0 1--124, 2021.

\bibitem[Huang et~al.(2017)Huang, Liu, Van Der~Maaten, and Weinberger]{huang2017densenet}
Gao Huang, Zhuang Liu, Laurens Van Der~Maaten, and Kilian~Q Weinberger.
\newblock Densely connected convolutional networks.
\newblock In \emph{Proceedings of the IEEE conference on computer vision and pattern recognition}, pages 4700--4708, 2017.

\bibitem[Kim et~al.(2017)Kim, Ahn, and Yoo]{kim2017zena}
Dongyoung Kim, Junwhan Ahn, and Sungjoo Yoo.
\newblock Zena: Zero-aware neural network accelerator.
\newblock \emph{IEEE Design \& Test}, 35\penalty0 (1):\penalty0 39--46, 2017.

\bibitem[Krizhevsky et~al.(2012)Krizhevsky, Sutskever, and Hinton]{krizhevsky2012imagenet}
Alex Krizhevsky, Ilya Sutskever, and Geoffrey~E Hinton.
\newblock Imagenet classification with deep convolutional neural networks.
\newblock \emph{Advances in neural information processing systems}, 25, 2012.

\bibitem[Kurtz et~al.(2020)Kurtz, Kopinsky, Gelashvili, Matveev, Carr, Goin, Leiserson, Moore, Shavit, and Alistarh]{kurtz2020inducing}
Mark Kurtz, Justin Kopinsky, Rati Gelashvili, Alexander Matveev, John Carr, Michael Goin, William Leiserson, Sage Moore, Nir Shavit, and Dan Alistarh.
\newblock Inducing and exploiting activation sparsity for fast inference on deep neural networks.
\newblock In \emph{International Conference on Machine Learning}, pages 5533--5543. PMLR, 2020.

\bibitem[LeCun et~al.(1998)LeCun, Bottou, Bengio, and Haffner]{lecun1998gradient}
Yann LeCun, L{\'e}on Bottou, Yoshua Bengio, and Patrick Haffner.
\newblock Gradient-based learning applied to document recognition.
\newblock \emph{Proceedings of the IEEE}, 86\penalty0 (11):\penalty0 2278--2324, 1998.

\bibitem[Lee et~al.(2019)Lee, Ajanthan, and Torr]{lee2018snip}
Namhoon Lee, Thalaiyasingam Ajanthan, and Philip Torr.
\newblock Snip: Single-shot network pruning based on connection sensitivity.
\newblock In \emph{International Conference on Learning Representations}, 2019.
\newblock URL \url{https://openreview.net/forum?id=B1VZqjAcYX}.

\bibitem[Li et~al.(2022)Li, You, Bhojanapalli, Li, Rawat, Reddi, Ye, Chern, Yu, Guo, et~al.]{li2022lazy}
Zonglin Li, Chong You, Srinadh Bhojanapalli, Daliang Li, Ankit~Singh Rawat, Sashank~J Reddi, Ke~Ye, Felix Chern, Felix Yu, Ruiqi Guo, et~al.
\newblock The lazy neuron phenomenon: On emergence of activation sparsity in transformers.
\newblock \emph{arXiv preprint arXiv:2210.06313}, 2022.

\bibitem[Liu et~al.(2019)Liu, Sun, Zhou, Huang, and Darrell]{liu2018rethinking}
Zhuang Liu, Mingjie Sun, Tinghui Zhou, Gao Huang, and Trevor Darrell.
\newblock Rethinking the value of network pruning.
\newblock In \emph{International Conference on Learning Representations}, 2019.

\bibitem[Ma et~al.(2019)Ma, Miao, Niu, and Zhang]{ma2019transformed}
Rongrong Ma, Jianyu Miao, Lingfeng Niu, and Peng Zhang.
\newblock Transformed ℓ1 regularization for learning sparse deep neural networks.
\newblock \emph{Neural Networks}, 119:\penalty0 286--298, 2019.

\bibitem[Mukherji et~al.(2023)Mukherji, Sch{\"o}ne, Nazeer, Mayr, and Subramoney]{mukherji2023activity}
Rishav Mukherji, Mark Sch{\"o}ne, Khaleelulla~Khan Nazeer, Christian Mayr, and Anand Subramoney.
\newblock Activity sparsity complements weight sparsity for efficient rnn inference.
\newblock In \emph{Machine Learning with New Compute Paradigms}, 2023.
\newblock URL \url{https://openreview.net/forum?id=hL4zZlCvWv}.

\bibitem[Natarajan(1995)]{natarajan1995sparse}
Balas~Kausik Natarajan.
\newblock Sparse approximate solutions to linear systems.
\newblock \emph{SIAM journal on computing}, 24\penalty0 (2):\penalty0 227--234, 1995.

\bibitem[Oh et~al.(2021)Oh, So, Kim, and Yi]{oh2021exploiting}
Chanyoung Oh, Junhyuk So, Sumin Kim, and Youngmin Yi.
\newblock Exploiting activation sparsity for fast cnn inference on mobile gpus.
\newblock \emph{ACM Transactions on Embedded Computing Systems (TECS)}, 20\penalty0 (5s):\penalty0 1--25, 2021.

\bibitem[Parashar et~al.(2017)Parashar, Rhu, Mukkara, Puglielli, Venkatesan, Khailany, Emer, Keckler, and Dally]{parashar2017scnn}
Angshuman Parashar, Minsoo Rhu, Anurag Mukkara, Antonio Puglielli, Rangharajan Venkatesan, Brucek Khailany, Joel Emer, Stephen~W Keckler, and William~J Dally.
\newblock Scnn: An accelerator for compressed-sparse convolutional neural networks.
\newblock \emph{ACM SIGARCH computer architecture news}, 45\penalty0 (2):\penalty0 27--40, 2017.

\bibitem[Raihan and Aamodt(2020)]{raihan2020sparse}
Md~Aamir Raihan and Tor Aamodt.
\newblock Sparse weight activation training.
\newblock \emph{Advances in Neural Information Processing Systems}, 33:\penalty0 15625--15638, 2020.

\bibitem[Reagen et~al.(2016)Reagen, Whatmough, Adolf, Rama, Lee, Lee, Hern{\'a}ndez-Lobato, Wei, and Brooks]{reagen2016minerva}
Brandon Reagen, Paul Whatmough, Robert Adolf, Saketh Rama, Hyunkwang Lee, Sae~Kyu Lee, Jos{\'e}~Miguel Hern{\'a}ndez-Lobato, Gu-Yeon Wei, and David Brooks.
\newblock Minerva: Enabling low-power, highly-accurate deep neural network accelerators.
\newblock \emph{ACM SIGARCH Computer Architecture News}, 44\penalty0 (3):\penalty0 267--278, 2016.

\bibitem[Szegedy et~al.(2015)Szegedy, Liu, Jia, Sermanet, Reed, Anguelov, Erhan, Vanhoucke, and Rabinovich]{szegedy2015going}
Christian Szegedy, Wei Liu, Yangqing Jia, Pierre Sermanet, Scott Reed, Dragomir Anguelov, Dumitru Erhan, Vincent Vanhoucke, and Andrew Rabinovich.
\newblock Going deeper with convolutions.
\newblock In \emph{Proceedings of the IEEE conference on computer vision and pattern recognition}, pages 1--9, 2015.

\bibitem[Tanaka et~al.(2020)Tanaka, Kunin, Yamins, and Ganguli]{tanaka2020pruning}
Hidenori Tanaka, Daniel Kunin, Daniel~L Yamins, and Surya Ganguli.
\newblock Pruning neural networks without any data by iteratively conserving synaptic flow.
\newblock \emph{Advances in neural information processing systems}, 33:\penalty0 6377--6389, 2020.

\bibitem[Tang et~al.(2021)Tang, Wang, Xu, Deng, Xu, Tao, and Xu]{tang2021manifold}
Yehui Tang, Yunhe Wang, Yixing Xu, Yiping Deng, Chao Xu, Dacheng Tao, and Chang Xu.
\newblock Manifold regularized dynamic network pruning.
\newblock In \emph{Proceedings of the IEEE/CVF conference on computer vision and pattern recognition}, pages 5018--5028, 2021.

\bibitem[Tian et~al.(2020)Tian, Krishnan, and Isola]{tian2019contrastive}
Yonglong Tian, Dilip Krishnan, and Phillip Isola.
\newblock Contrastive representation distillation.
\newblock \emph{ICLR}, 2020.

\bibitem[Wang et~al.(2021)Wang, Zhang, Xie, Guo, Liu, and Leng]{wang2021dual}
Yang Wang, Chen Zhang, Zhiqiang Xie, Cong Guo, Yunxin Liu, and Jingwen Leng.
\newblock Dual-side sparse tensor core.
\newblock In \emph{2021 ACM/IEEE 48th Annual International Symposium on Computer Architecture (ISCA)}, pages 1083--1095. IEEE, 2021.

\bibitem[Yang et~al.(2019)Yang, Mao, Wang, and Li]{yang2019dasnet}
Qing Yang, Jiachen Mao, Zuoguan Wang, and Hai Li.
\newblock Dasnet: Dynamic activation sparsity for neural network efficiency improvement.
\newblock In \emph{2019 IEEE 31st International Conference on Tools with Artificial Intelligence (ICTAI)}, pages 1401--1405. IEEE, 2019.

\bibitem[Zhang(2017)]{zhang2017transformed}
Shuai Zhang.
\newblock \emph{Transformed ℓ1 Function, Sparse Optimization Algorithms and Applications}.
\newblock University of California, Irvine, 2017.

\bibitem[Zhu et~al.(2022)Zhu, Pourtaherian, Waeijen, Bamberg, Bondarev, and Moreira]{zhu2022arts}
Zeqi Zhu, Arash Pourtaherian, Luc Waeijen, Lennart Bamberg, Egor Bondarev, and Orlando Moreira.
\newblock Arts: An adaptive regularization training schedule for activation sparsity exploration.
\newblock In \emph{2022 25th Euromicro Conference on Digital System Design (DSD)}, pages 415--422. IEEE, 2022.

\bibitem[Zhu et~al.(2023)Zhu, Pourtaherian, Waeijen, Bondarev, and Moreira]{zhu2023star}
Zeqi Zhu, Arash Pourtaherian, Luc Waeijen, Egor Bondarev, and Orlando Moreira.
\newblock Star: Sparse thresholded activation under partial-regularization for activation sparsity exploration.
\newblock In \emph{2023 IEEE/CVF Conference on Computer Vision and Pattern Recognition Workshops (CVPRW)}, pages 4554--4563. IEEE, 2023.

\end{thebibliography}
\bibliographystyle{plainnat}

\newpage
\section*{Supplementary Material}
In the supplementary material, We present the regularization parameters and tuning parameters per layer for each network in Tables \ref{table:paramForTL1},\ref{table:paramForDual}. \cite{zhu2022arts} developed an adaptive method to select appropriate regularization values for each layer, this requires a significant time cost. On the CIFAR-10 dataset, selecting suitable values for ResNet18 requires 850 epochs. Additionally, Transformed $\ell1$ involves a tuning parameter. We set a global regularization parameter and a tuning parameter for all models. 

\begin{table}[ht]
    \caption{Regularization parameters and tuning parameters for LeNet5\citep{lecun1998gradient}, GoogleNet\citep{szegedy2015going}, DenseNet121\citep{huang2017densenet} and ResNet18/34/50\citep{he2016resnet}, in activation map sparsity induction experiment.}\label{table:paramForTL1}
    \centering
    \begin{tabular}{lll}
    \toprule
    \textbf{Model} & $\alpha_l(l=1,...,L-1)$  & $\beta_l(l=1,...,L-1)$\\
    \midrule
        LeNet5      & $1.0\times10^{-8}$  & $1.0\times10^{-4}$\\
        GoogLeNet   & $5.0\times10^{-10}$ & $1.0\times10^{-3}$\\
        DenseNet121 & $5.0\times10^{-11}$ & $1.0\times10^{-4}$\\
        ResNet18    & $2.0\times10^{-9}$  & $1.0\times10^{-2}$\\
        ResNet34    & $1.0\times10^{-9}$  & $1.0\times10^{-2}$\\
        ResNet50    & $4.0\times10^{-10}$ & $1.0\times10^{-2}$\\
    \bottomrule
    \end{tabular}
\end{table}

\begin{table}[ht]
    \caption{Regularization parameters and tuning parameters for ResNet18/34/50\citep{he2016resnet}, in dual sparse training experiment.}\label{table:paramForDual}
    \centering
    \begin{tabular}{lll}
    \toprule
    \textbf{Model} & $\alpha_l(l=1,...,L-1)$  & $\beta_l(l=1,...,L-1)$\\
    \midrule
        ResNet18    & $3.0\times10^{-9}$  & $1.0\times10^{-2}$\\
        ResNet34    & $1.0\times10^{-9}$  & $1.0\times10^{-2}$\\
        ResNet50    & $5.0\times10^{-10}$ & $1.0\times10^{-2}$\\
    \bottomrule
    \end{tabular}
\end{table}
\end{document}